\documentclass{SCIS2019}

\begin{document}
\ArticleType{RESEARCH PAPER}
\Year{2019}
\Month{}
\Vol{}
\No{}
\DOI{}
\ArtNo{}
\ReceiveDate{}
\ReviseDate{}
\AcceptDate{}
\OnlineDate{}

\title{FACLSTM: ConvLSTM with Focused Attention for Scene Text Recognition}{FACLSTM: ConvLSTM with Focused Attention for Scene Text Recognition}

\author[1,2]{Qingqing Wang}{}
\author[2]{Ye Huang}{}
\author[2]{Wenjing Jia}{}
\author[2]{Xiangjian He}{}
\author[2]{Michael Blumenstein}{}
\author[1]{\\Shujing Lyu}{}

\author[1,3]{Yue Lu}{{ylu@cs.ecnu.edu.cn}}
\AuthorMark{Yue Lu}

\AuthorCitation{Qingqing Wang, Ye Huang, Wenjing Jia, et al}


\address[1]{Shanghai Key Laboratory of Multidimensional Information Processing, East China Normal University, Shanghai {\rm 200241}, China}
\address[2]{Faculty of Engineering and Information Technology, University of Technology Sydney, Sydney {\rm 2007}, Australia}
\address[3]{Shanghai Institute of Intelligent Science and Technology, Tongji University, Shanghai {\rm 200092}, China}

\abstract{Scene text recognition has recently been widely treated as a sequence-to-sequence prediction problem, where traditional fully-connected-LSTM (FC-LSTM) has played a critical role. 
Due to the limitation of FC-LSTM, existing methods have to convert 2-D feature maps into 1-D sequential feature vectors, resulting in severe damages of the valuable spatial and structural information of text images.
In this paper, we argue that scene text recognition is essentially a spatiotemporal prediction problem for its 2-D image inputs, and propose a convolution LSTM (ConvLSTM)-based scene text recognizer, namely, FACLSTM, \textit{i.e.,} {\color{red}F}ocused {\color{red}A}ttention {\color{red}C}onv{\color{red}LSTM}, where the spatial correlation of pixels is fully leveraged when performing sequential prediction with LSTM.
Particularly, the attention mechanism is properly incorporated into an efficient ConvLSTM structure via the convolutional operations and additional character center masks are generated to help focus attention on right feature areas.   
The experimental results on benchmark datasets IIIT5K, SVT and CUTE demonstrate that our proposed FACLSTM performs competitively on the regular, low-resolution and noisy text images, and outperforms the state-of-the-art approaches on the curved text images with large margins.}


\maketitle

\section{Introduction}
\begin{figure}[t]
\centering
\includegraphics[width=2.4in]{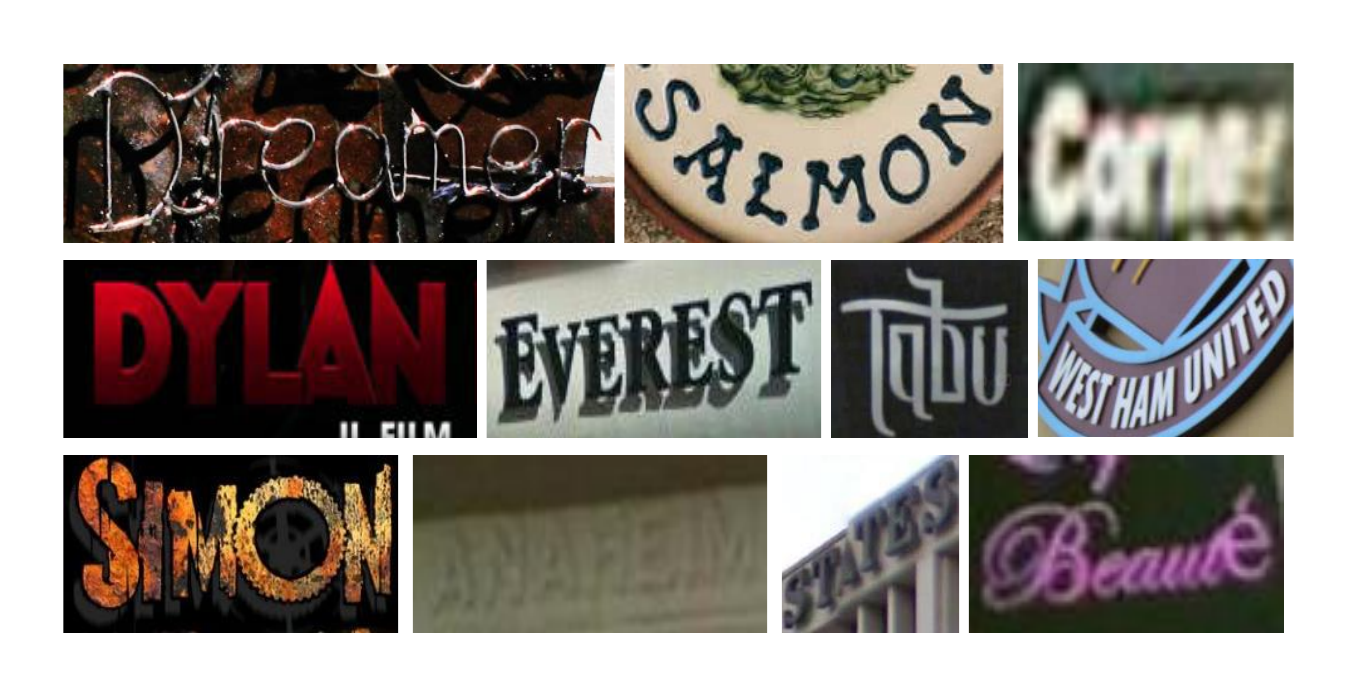}
\caption{Challenging samples of scene text recognition.}
\label{Figure0}
\end{figure}

Scene text recognition has received considerable attention from the community of computer vision since text is an essential way to convey information and knowledge. Due to the challenges posed by poor image qualities (\textit{e.g.}, low resolution, blur, and uneven illumination) and various text appearances (\textit{e.g.}, size, fonts, colors, directions, perspective view as well as complex background), as shown in Fig.~\ref{Figure0}, though many efforts have been made in past decades, scene text recognition is still an unsolved task.

\begin{figure}[t]
\centering
\includegraphics[width=4in]{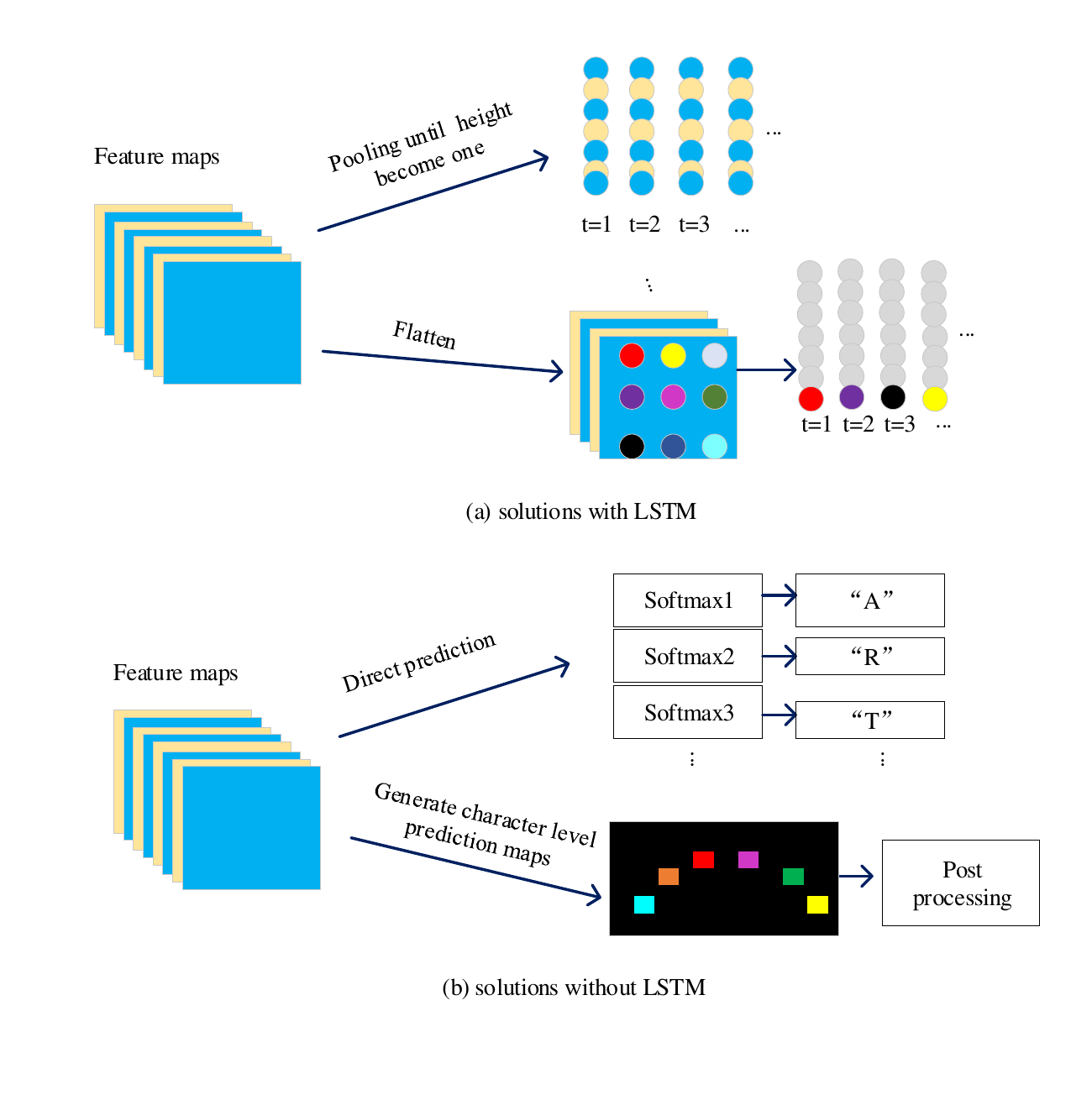}
\caption{Current solutions for scene text recognition. When using LSTM, 2-D feature maps are usually converted to 1-D space by pooling or flattening operations. When the LSTM is not used, additional parameters or post precessing steps are involved.}
\label{Figure1}
\end{figure}

Inspired by speech recognition and machine translation, most of recent state-of-the-art approaches regard scene text recognition as a sequence-to-sequence prediction problem and widely adopt techniques like LSTM~\cite{ref28} and attention mechanism~\cite{ref17,ref18} in their sequential transcription module.
However, the LSTM used in these recognizers is the fully-connected-LSTM (FC-LSTM) that only takes stream signals like sentences or audio as inputs and connects them in a fully connected way, while scene text recognition generates sequential outputs from 2-D images.
To adapt FC-LSTM to scene text recognition, the most straightforward way is pooling 2-D feature maps to a height of one or flattening them into 1-D sequential feature vectors~\cite{ref2, ref4, ref5, ref6, ref7}, as shown in Fig~\ref{Figure1}(a). 
Unfortunately, such operations could severely disrupt the valuable spatial correlation relationships among pixels, which is essential to computer vision tasks, especially to scene text recognition, where the structures of strokes are the key factors to discriminate characters.   
To retain such important spatial and structural information, researchers have also explored other alternative solutions. For example,
STN-OCR~\cite{ref8} directly performed sequential prediction on 2-D feature maps with a fixed number of softmax classifiers; CA-FCN~\cite{ref3} generated character-level confidence maps with a fully convolutional network, as shown in Fig.~\ref{Figure1}(b). 
However, compared with LSTM, these solutions often introduce additional parameters or post processing steps.  

In this paper, we propose to address the issue of scene text recognition from the perspective of spatiotemporal prediction, where the spatial correlation information is taken into account when performing sequential prediction with LSTM. 
The ConvLSTM proposed by Shi et al.~\cite{ref31} for precipitation nowcasting provides some insights on how to achieve this. In ConvLSTM, all of the fully connected operations are replaced by convolutional ones, so input feature maps are allowed to keep their 2-D shape when being fed into the ConvLSTM. Given this advantage, for the first time, we introduce ConvLSTM to scene text recognition and apply it in the sequential transcription module of our proposed recognizer.   

However, in existing models, both FC-LSTM and ConvLSTM are used only for frame-level prediction and are incapable of producing sequential outputs from one single input image unless the Connectionist Temporal Classification (CTC)~\cite{ref1,ref2,ref6} or attention mechanism~\cite{ref4, ref5, ref7, ref12} is incorporated. 
To perform sequential prediction and, meanwhile, provide the model spatial awareness, we further improve ConvLSTM by embedding the attention mechanism into the structure. Notably, different from the existing attention-LSTM-based recognizers, where the attention mechanism and FC-LSTM are combined in a fully connected way, we properly integrate the attention mechanism into ConvLSTM with the convolutional operations. 
Moreover, as ConvLSTM extends 2-D operations into 3-D, the costs of computation and memory increase significantly. To achieve high efficiency, inspired by Liu et al.~\cite{ref27}, we propose to assemble a bottleneck gate at the beginning of the proposed attention-equipped ConvLSTM, so that the internal feature map channels can be reduced.
 
Last but not the least, since existing attention-based recognizers often suffer from the `attention drift' problem~\cite{ref4}, \textit{i.e.}, they fail to align target outputs to proper feature areas, we propose to learn additional character center masks with a second decoder branch in the encoder-decoder feature extraction stage to assist the proposed network to focus attention on right feature areas.
The experimental results conducted on benchmark datasets demonstrate that our proposed recognizer is able to achieve comparable performance with the state-of-the-art approaches on regular, low-resolution and noisy text and outperforms other methods significantly on the more challenging curved text.

The contributions made in this work are summarized as follows. 
(1) We propose to handle the scene text recognition problem from a spatiotemporal prediction perspective and for the first time introduce ConvLSTM to this application. 
(2) We design a ConvLSTM-based sequential transcription module, where the attention mechanism is harmoniously embedded into ConvLSTM with convolutional operations, and the bottleneck gate is assembled at the beginning of ConvLSTM to retain its efficiency. 
(3) We propose to learn additional character center masks to help the proposed network focus attention on the center of characters. 

In the rest of this paper, we first review the most related works in Section~\ref{sec:relatedWork}. 
Then, the details of our proposed approach and designed experiments are presented in Sections~\ref{sec:Method} and~\ref{sec:exp}, respectively. Finally, the conclusions are given in Section~\ref{sec:con}. 

\section{Related works}
\label{sec:relatedWork}

The existing scene text recognizers can be grouped into two categories, \textit{i.e.}, the ones utilizing traditional techniques and the ones based on deep learning techniques. 
Methods belonging to the first category were mainly proposed before 2015, and follow a bottom-up routine, \textit{i.e.}, detecting and recognizing individual characters first, followed by word formation. 
Ye et al.~\cite{ref15} provided a comprehensive survey for these methods. 
By contrast, the deep learning-based recognizers depend on end-to-end trainable deep networks, where feature extraction and sequential translation are integrated into one unified framework. 
According to literature, the deep learning-based recognizers are now the dominant solutions to scene text recognition, and surpass traditional ones by large margins. 
Therefore, in this section, we only review recognizers applying deep learning techniques, along with ConvLSTM and related variants.

\textbf{Methods based on LSTM:}
LSTM is widely used in the existing state-of-the-art recognizers for three purposes, \textit{i.e.,} producing frame-level predictions required by the subsequent sequential transcription module~\cite{ref2, ref6}, encoding sequential features with considering historical information~\cite{ref7,ref40}, and directly generating sequential predictions when cooperating with the attention mechanism~\cite{ref4, ref5, ref12, ref25, ref40}.
For example, CRNN proposed by Shi et al.~\cite{ref6} was composed of three parts, \textit{i.e.,} the convolution module used to extract features from input images, a bi-LSTM layer built to make predictions for individual frames, and a CTC-based sequential transcription component utilized to infer sequential outputs from frame-level predictions.
As clarified in~\cite{ref41}, irregularly shaped art text presents frequently in our daily life, especially perspective text and curved text, which have posed enormous challenges for scene text recognition. To tackle this problem,
Shi et al.~\cite{ref7} employed a bi-LSTM layer in their RARE to extract sequential feature vectors from input feature maps, followed by feeding these vectors into an attention-Gated Recurrent Unit (GRU) module to generate label sequences. A highlight of RARE was its usage of Spatial Transformer Network (STN)~\cite{ref32}, which was responsible for rectifying images containing irregular texts and was widely adopted by subsequent recognizers like STN-OCR~\cite{ref8} and ASTER~\cite{ref40}. 
Afterwards, RARE was extended to ASTER~\cite{ref40} by modifying the architecture of rectification network. Note that, LSTM was used for both feature encoding and sequential transcription in ASTER. 
Lee et al.~\cite{ref25} combined a recursive CNN with a recurrent CNN in their R$^2$ AM to capture long-term dependencies when extracting features from raw images, and then fed these features to an attention-RNN network for sequential transcription.
Gao et al.~\cite{ref1, ref2} designed two models to compare the performance of CNN and LSTM in terms of sequential feature encoding. According to their experiments, features extracted by LSTM were more powerful than those extracted by CNN. 
Cheng et al.~\cite{ref4, ref5} combined LSTM with an attention mechanism in the sequential transcription module of their FAN and AON recognizers, but they criticized that the existing attention-based models often failed to align attention to right feature areas when performing prediction. 
Therefore, a focusing network was assembled in their FAN~\cite{ref4} to tackle this problem. 
AON~\cite{ref5} was specially designed for irregular text recognition. In this work, features were extracted from four directions, and then combined and filtered with a filter gate.
Wojna et al.~\cite{ref12} utilized an attention-equipped LSTM to localize and recognize text from street view images. Their model was given location awareness by incorporating one-hot encoded spatial coordinates into the LSTM.
Bai et al.~\cite{ref43} pointed out that exiting attention-based recognizers failed to align ground truth strings with attention's probability outputs, and this confused and misled the training process of the networks. To tackle this problem, they proposed Edit Probability (EP), which took the possible occurrences of missing and superfluous characters into consideration when estimating the probability of generating a string from the network's outputs. Su et al.~\cite{ref44, ref46} converted text images into sequential signals via extracting their HOG features, and designed an ensembling technique to combine the outputs of two LSTM branches, so that better recognition performance could be achieved.
Li et al~\cite{ref47} pointed out that traditional attention mechanism was not able to produce accurate attention predictions, thus the recognition performance on irregular text images was largely compromised. To address this issue, they designed a 2-D attention module, where one LSTM was used to encode feature maps column by column to produce holistic features, and another was employed as usual to generate final sequential outputs.
Note that, the LSTM used in all the methods mentioned above refers to traditional FC-LSTM, so the 2-D feature maps have to be mapped into 1-D space in order to adapt to the LSTM layers, and the attention mechanism has to be incorporated in a fully connected way. This severely damages the spatial and structural information of input images, which is essential to computer vision tasks such as scene text recognition.

\textbf{Methods without LSTM:} 
At the beginning of the deep learning era, a group of Deep CNN (DCNN) recognizers~\cite{ref16, ref17} were well developed and made breakthrough over traditional recognizers. For instance, Tian et al.~\cite{ref45} proposed two feature descriptors, \textit{i.e.,} Co-occurrence HOG (Co-HOG) and Convolutional Co-HOG, and combined them with CNN to perform scene text recognition on multiple languages.
In these models, CNN together with softmax classifier were widely used for character or word classification. However, with the development of LSTM-based recognizers, the DCNN ones were quickly and significantly surpassed. 
Recently, some researchers argued that LSTM-based models were hard to train~\cite{ref1} and not able to achieve good performance on non-horizontal text~\cite{ref3}, so explorations on models without LSTM started again. 
For instance, STN-OCR~\cite{ref8} utilized fully connected layers and a fixed number of softmax classifiers for sequential prediction;
SqueezedText~\cite{ref9} employed a binary convolutional encoder-decoder network to generate salience maps for individual characters and then exploited a GRU-based bi-RNN for further correction;
Liao et al.~\cite{ref3} proposed to address the scene text recognition issue from a 2-D perspective with a CA-FCN model, so that the spatial information could be taken into account when performing prediction.
As proved in~\cite{ref42}, the performance of object recognition has been largely fueled by the detection of salience regions. Therefore, in CA-FCN, a character attention module was utilized to produce pixel-level confidence maps for target characters, and then these maps were fed into a word formation module to generate word-level outputs.

Different from those LSTM-based approaches, recognizers without LSTM can better leverage the spatial information, but they also unavoidably introduce additional parameters or post processing steps in order to produce sequential outputs, such as the multiple classifiers used by STN-OCR~\cite{ref8} and the word formation module designed in~\cite{ref3}. 

\begin{figure}[t]
\centering
\includegraphics[width=5.8in]{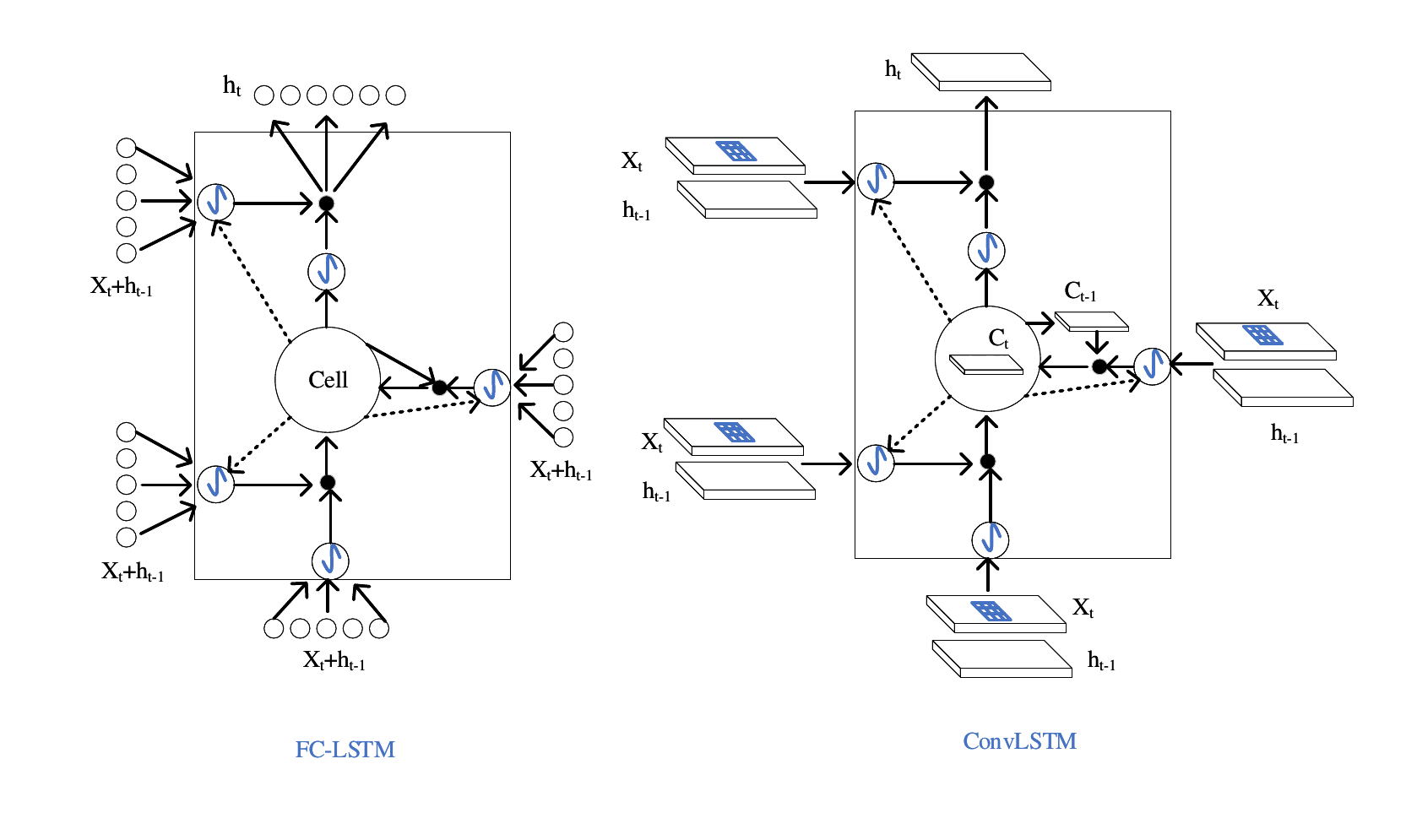}
\caption{Illustration of the FC-LSTM (left) and the ConvLSTM (right). The FC-LSTM is performed in 1-D space, while the ConvLSTM is performed in 2-D space.} 
\label{Figure2}
\end{figure}

\textbf{ConvLSTM and Related Variants:}
As explained in~\cite{ref31}, the main drawback of traditional FC-LSTM was its usage of full connections in the input-to-state and state-to-state transitions, which resulted in the neglect of spatial information. To retain such important information, ConvLSTM, proposed by Shi et al.~\cite{ref31}, replaced all of the full connections of traditional FC-LSTM with convolutional operations, and extended the 2-D features and states into 3-D, as shown in Fig.~\ref{Figure2}.
Their experimental results demonstrated the superiority of ConvLSTM over traditional FC-LSTM. 
Thereafter, some variants of ConvLSTM have been developed for action recognition~\cite{ref33}, object detection in video~\cite{ref27}, and gesture recognition~\cite{ref35, ref34} etc. 
For example, Zhu et al.~\cite{ref34} combined ConvLSTM with the 3-D convolution in a multimodal model, and achieved promising gesture recognition performance.
Li et al.~\cite{ref33} designed a motion-based attention mechanism and combined it with ConvLSTM in their VideoLSTM, which is proposed for action recognition in videos.

In our work, aiming to better consider the spatial and structural information of input images when performing sequential prediction with LSTM, for the first time, we propose an attention-equipped ConvLSTM structure in the sequential transcription module, and further design a focused attention module to help learn more accurate alignment between predicted characters and corresponding feature areas.

\section{Methodology} 
\label{sec:Method}
\begin{figure*}[t]
\centering
\includegraphics[width=6.4in]{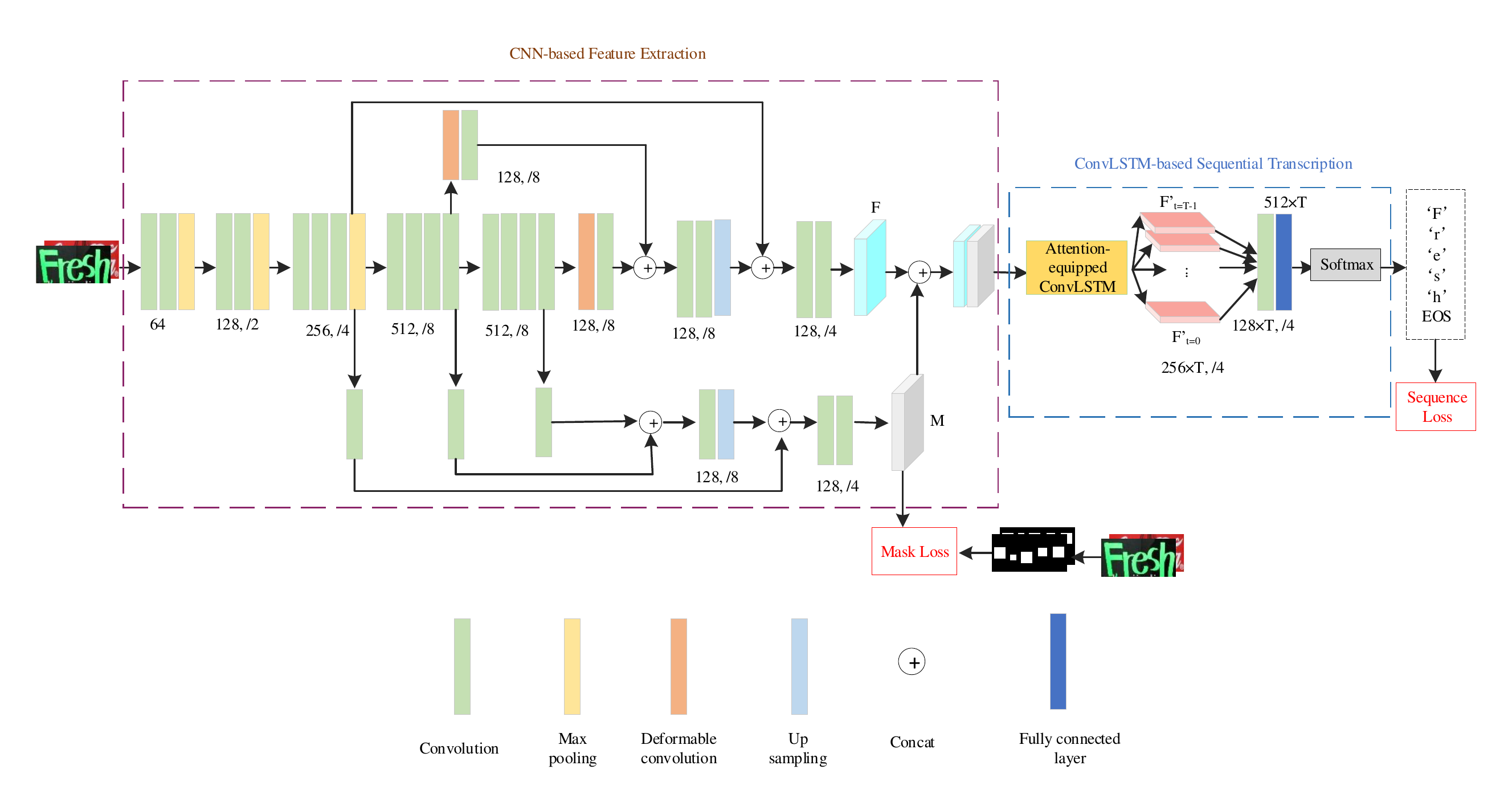}
\caption{Overview of proposed FACLSTM. $F$ and $M$ denote the extracted feature maps and character center masks. $T$ groups of feature maps are produced by the proposed attention-eqipped ConvLSTM, where $T$ is the maximal string length, and the followed softmax classifier is responsible for producing $T$ groups of feature maps from extracted feature maps. Note that, the softmax classifier and previous fully connected layer are shared by the $T$ groups of feature maps.}
\label{Figure3}
\end{figure*}

As illustrated in Fig.~\ref{Figure3}, our proposed FACLSTM, \textit{i.e.,} Focused Attention ConvLSTM, consists of two components, \textit{i.e.}, the CNN-based feature extraction module and the ConvLSTM-based sequential transcription module. 
The feature extraction module is an encoder-decoder structure that takes VGG-16 as the backbone, while the sequential transcription module is a combination of ConvLSTM and attention mechanism. More details are presented as follows. 

\subsection{CNN-based Feature Extraction}

\textbf{Backbone:} Similar to Liao's work~\cite{ref3}, we take VGG-16 as the encoder of our feature extraction module, and remove the fully connected layers and pooling layers from the last two encoding stages. We also assemble two deformable convolutional layers~\cite{ref36} at stage-4 and stage-5 of the decoder given their flexible receptive fields.
However, compared with Liao's network~\cite{ref3}, the resolution of final feature maps is restored to a smaller size of $\frac{W}{4}\times \frac{H}{4} \times C$ in our FACLSTM, instead of the $\frac{W}{2}\times \frac{H}{2} \times C$ used in~\cite{ref3}, considering the memory and computation cost. Here, $W$, $H$ and $C$ denote the width, height and channels of feature maps, respectively. In addition, we remove their character attention module set in the encoder stage, and meanwhile, design a focused attention module in the higher-level decoder stage so that more abstract and powerful character center masks can be extracted.  

\textbf{Focused Attention Module:} 
As pointed out in~\cite{ref4}, current attention-based models suffer from the `attention drift' problem, \textit{i.e.,} they fail to obtain accurate alignment between target characters and related feature areas, especially in complicated and low-quality images. 
To tackle this problem, in the feature extraction module of the proposed FACLSTM, we assemble two decoder branches, of which one is used as normal for feature extraction and another is designed to learn additional character center masks as centers of text regions are always the key to scene text detection~\cite{ref48} and recognition~\cite{ref3}. 
These masks are expected to guide the subsequent attention module regarding where to focus. Obviously, for each timestep, the attention should be focused on the center of certain character. Moreover, these masks can also help to enhance foreground pixels and suppress background pixels.  

In other recognizers~\cite{ref1,ref2,ref3}, the feature maps $F$ and maps $A$ generated for other purposes are always combined with the element-wise multiplication $\otimes$ in the way of $F_{out}=F\otimes(1+A)$.
However, in our experiments we find that directly concatenating feature maps $F$ and character center masks $M$ achieves better performance, which means the subsequent attention-based module prefers to learn patterns from $F$ and $M$ directly, rather than from their fused results. Therefore, direct concatenation $F_{out}=F\oplus M$ is used in our FACLSTM. 

\subsection{Sequential Transcription Module}

As shown in Fig.~\ref{Figure3}, our sequential transcription module starts with an attention-equipped ConvLSTM, by which $T$ groups of feature maps with the size of $\frac{W}{4}\times \frac{H}{4} \times C$ are generated. Here, $T$ is the predefined maximal string length.  
Afterwards, a $1\times 1$ convolutional layer is applied to reduce the feature map channels, followed by a fully connected layer and a softmax classifier that are employed to sequentially predict $T$ characters. 
Details of proposed sequential transcription module are presented below.

\textbf{ConvLSTM:} The structure of the traditional FC-LSTM~\cite{ref28} is illustrated in Fig.~\ref{Figure2}(left), and related key formulations can be expressed as Eq.~\ref{equ1}, where $\circ$ is the Hadamard product (\textit{i.e.}, element-wise multiplication), $f$ denotes the activation function of input gate $i_t$, output gate $o_t$ and forget gate $f_t$, and $x_t$, $c_t$ and $h_t$ represent input features, cell states and cell outputs, respectively.  
\begin{equation}
\begin{array}{lr}
i_t =f(w_{xi}x_t+w_{hi}h_{t-1}+w_{ci}\circ c_{t-1}) &\\
f_t = f(w_{xf}x_t+w_{hf}h_{t-1}+w_{cf}\circ c_{t-1}) &\\
c_t =f_t\circ c_{t-1}+i_t\circ tanh(w_{xc}x_t+w_{hc}h_{t-1}) &\\
o_t =f(w_{xo}x_t+w_{ho}h_{t-1}+w_{co}\circ c_t) &\\
h_t =o_t\circ tanh(c_t).
\end{array}
\label{equ1}
\end{equation}

As we can see, FC-LSTM takes 1-D sequential feature vectors as input, and calculates both the input-to-state and state-to-state transactions in a fully connected manner.
Therefore, when applying it to computer vision tasks, the 2-D feature maps have to be mapped into 1-D space, during which the spatial correlation relationships among pixels are badly damaged.  

To take advantage of such valuable spatial and structural information in computer vision tasks, Shi et al.~\cite{ref31} proposed ConvLSTM by incorporating convolutional structures into LSTM. 
As shown in Fig.~\ref{Figure2}(right), all input features, gates, cell states and cell outputs are 3-D in ConvLSTM, and all of the input-to-state and state-to-state transactions are performed with the convolutional operations, instead of the fully connected ones. 
Thus, the key formulations of ConvLSTM can be written as Eq.~\ref{equ2}, where $*$ denotes the convolutional operation.
\begin{equation}
\begin{array}{lr}
i_t =f(w_{xi}*x_t+w_{hi}*h_{t-1}+w_{ci}\circ c_{t-1}) &\\
f_t = f(w_{xf}*x_t+w_{hf}*h_{t-1}+w_{cf}\circ c_{t-1}) &\\
c_t =f_t\circ c_{t-1}+i_t\circ tanh(w_{xc}*x_t+w_{hc}*h_{t-1})  &\\
o_t =f(w_{xo}*x_t+w_{ho}*h_{t-1}+w_{co}\circ c_t) &\\
h_t =o_t\circ tanh(c_t).
\end{array}
\label{equ2}
\end{equation}

\textbf{Proposed Attention-equipped ConvLSTM:}
The attention mechanism has achieved excellent performance in sequential prediction tasks, such as machine translation~\cite{ref17}, speech recognition~\cite{ref18}, as well as scene text recognition~\cite{ref4, ref5, ref12, ref25, ref40}. 
Especially, in the field of scene text recognition, it has been widely combined with FC-LSTM or GRU to produce more accurate predictions.
On the other hand, LSTM is used only for frame-level prediction in the existing works and is seldom utilized for producing sequential outputs from one single input image unless when combined with the CTC or attention mechanism.
\begin{figure*}[t]
\centering
\includegraphics[width=5.8in]{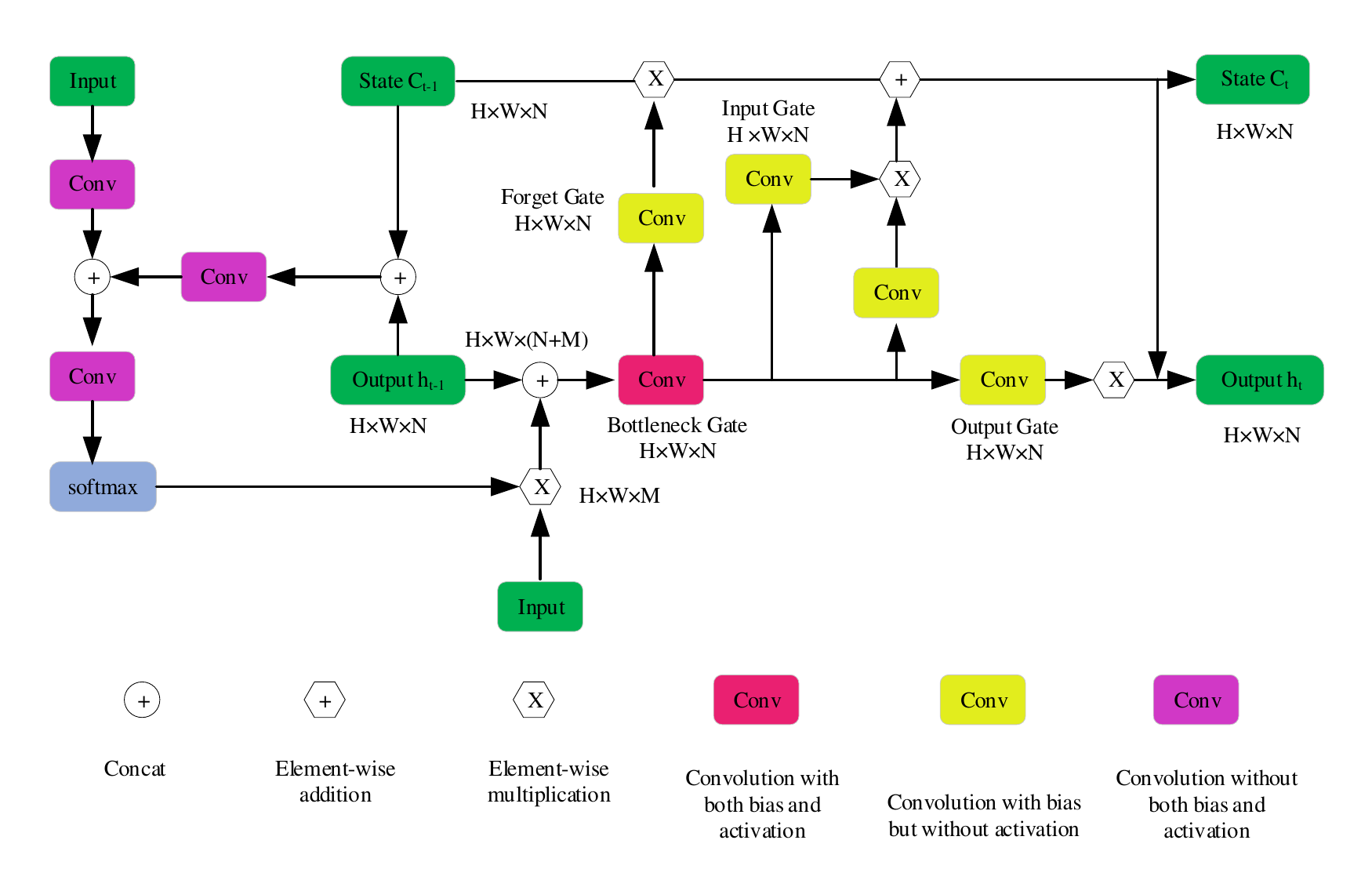}
\caption{Illustration of our proposed attention-equipped ConvLSTM, where the inputs are weighted by attention scores derived from previous cell state and cell output.}
\label{Figure5}
\end{figure*}

Therefore, in this work, to adapt ConvLSTM to scene text recognition and, meanwhile, provide the proposed network location awareness, we incorporate the attention mechanism into ConvLSTM by weighting the input feature maps with attention scores derived from the cell states and cell outputs obtained at the previous timestep, as illustrated in Fig.~\ref{Figure5}. In addition, to retain the efficiency of the proposed network, an additional bottleneck gate is assembled before the original input gate, forget gate and output gate to reduce the internal feature map channels.

Eqs.~\ref{equ4} and~\ref{equ5} provide more details on how the cell outputs and the attention scores are calculated. Here, $[\cdot,\cdot]$ is the channel-wise concatenation, $R(\cdot)$ and $S(\cdot)$ denote the ReLU activation function and the Sigmoid function, respectively, and $\widehat{x}_t$ represents the weighted inputs computed by Eq.~\ref{equ5}.
Keep it in mind that all of the gates $\{b,i,o,f\}_t$, inputs $\widehat{x}_t$, cell states $c_{\{t,t-1\}}$ and cell outputs $h_{\{t,t-1\}}$ in Eqs.~\ref{equ4} and~\ref{equ5} are in 3-D. Moreover, $w_{\{b,i,f,o,b2,h,x\}}$ and $bias_{\{b,i,f,o,f2,b2,y\}}$ are the involved network weights and biases, and $x_t$ is the concatenation of feature maps $F$ and character center masks $M$ produced by aforementioned encoder-decoder feature extraction module.

\begin{equation}
\begin{array}{lr}
b_t =R(w_b*([\widehat{x}_t, h_{t-1}])+bias_b) &\\
i_t =w_i*b_t+bias_i &\\
f_t =w_f*b_t+bias_f &\\
o_t =w_o*b_t+bias_o &\\
c_t =S(f_t+bias_{f2})\circ c_{t-1}+S(i_t)\circ R(w_{b2}*b_t+bias_{b2}) &\\
h_t =R(c_t)\circ S(o_t) 
\end{array}
\label{equ4}
\end{equation}

\begin{equation}
\begin{array}{lr}
h_{yt} =[w_h*[c_{t-1}, h_{t-1}], (w_x*x)]+bias_y &\\ 
z_t =w_z*tanh(h_{yt}) &\\
attn_t =softmax(z_t) &\\
\widehat{x}_t =attn_t \circ x 
\end{array}
\label{equ5}
\end{equation}

Once the cell outputs $H=\{h_1, h_2, ..., h_T\}, h_i\in R^{M\times N \times C}$ are obtained from the proposed attention-equipped ConvLSTM, a $1\times 1$ convolutional layer is applied to map them to $\widetilde{H}=\{\widetilde{h}_1, \widetilde{h}_2, ..., \widetilde{h}_T\}, \widetilde{h}_i\in R^{M\times N \times \widetilde{C}}$ and $\widetilde{C}<C$, which is also used to improve model's efficiency, just like the bottleneck gate does. 
Afterwards, a fully connected layer and a softmax classifier are designed to generate the final sequential outputs $S=\{c_1, c_2, ..., c_T\}$ from $\widetilde{H}$, where $c_i$ is from the predefined charset. Compared with STN-OCR~\cite{ref8}, where multiple fully connected layers and multiple softmax classifiers are assembled for sequential transcription, in our FACLSTM, only one single fully connected layer and one softmax classifier are employed and shared by $T$ groups of feature maps.

\subsection{Training}
\textbf{Loss function:} The objective function $L$ of our proposed FACLSTM consists of two parts, \textit{i.e.}, the sequential prediction loss $L_s$ and the mask loss $L_m$, as formulated in Eq.~\ref{equ6}, where $m$, $\widetilde{m}$, $\widehat{y}$ and $\widetilde{y}$ are the ground truth masks, predicted masks, smoothed ground truth strings and predicted sequential outputs, respectively. 
$\lambda$ is the coefficient used to balance the importance of the sequential prediction loss and the mask loss, and is set to $1$ in our experiments. 
Additionally, the label smoothing method proposed by Szegedy et al.~\cite{ref19} is able to help regularize the proposed model. Therefore, given the one-hot encoded ground truth $y^{OneHot}$, we convert it to the smoothed version $\widehat{y}$ with Eq.~\ref{equ7}. 
Moreover, for the ground truth masks $m$, we set the value of their foreground pixels (center of characters) and background pixels to 1 and 0, respectively. Thus, the mask loss $Lm$ is calculated in the way of Eq.~\ref{equ8}.
\begin{equation}
L=Ls(\widehat{y}, \widetilde{y})+\lambda Lm(m, \widetilde{m})
\label{equ6}
\end{equation}
\begin{equation}
\widehat{y}=(1.0-\epsilon)*y^{OneHot}+\epsilon*(\frac{1}{N_{class}})
\label{equ7}
\end{equation}
\begin{equation}
Lm=0.01*\{1-2*[\frac{\sum (m\otimes\widetilde{m})}{\sum m +\sum \widetilde{m}}]\}
\label{equ8}
\end{equation}

\textbf{Generation of Ground Truth:} 
Obviously, to optimize the proposed network, ground truth of character center masks is required.
Assuming $b=(x_{min}, y_{min}, x_{max}, y_{max})$ is the bounding box of individual characters, we use the same method as that in~\cite{ref3} to calculate the ground truth of the corresponding mask $g=(x_{min}^g, y_{min}^g, x_{max}^g, y_{max}^g)$, as shown in Eq.~\ref{equ9}. 
\begin{equation}
\begin{array}{lr}
w =x_{max}-x_{min} &\\
h =y_{max}-y_{min} &\\
x_{min}^g =(x_min+x_max-w*r)/2 &\\
x_{max}^g =(x_min+x_max+w*r)/2 &\\
y_{min}^g =(y_min+y_max-h*r)/2 &\\
y_{max}^g =(y_min+y_max+h*r)/2 &\\
\end{array}
\label{equ9}
\end{equation} 
Note that, the shrink ratio $r$ is set to 0.25 in our experiments, instead of 0.5 used in~\cite{ref3}.    

\section{Experiments}
\label{sec:exp}
\subsection{Datasets}

We train the proposed FACLSTM network with 7 million synthetic images from SynthText dataset~\cite{ref20} (available at \url{http://www.robots.ox.ac.uk/~vgg/data/scenetext/}) without fine-tuning on individual real-word datasets, and evaluate the corresponding performance on three widely used benchmarks, including the regular text dataset IIIT5K, low-resolution and noisy text dataset SVT, and curved text dataset CUTE.

\begin{itemize} 
\item \textbf{SynthText} is proposed by Gupta et al.~\cite{ref20} for scene text detection. The original dataset is composed of 800,000 scene text images, each with multiple word instances. Texts in this dataset are rendered in different styles, and annotated with character-level bounding boxes. Overall, about 7 million text images are cropped for scene text recognition. 
\item \textbf{IIIT5K} is built by Mishra et al.~\cite{ref22}. This dataset consists of 3000 text images obtained from the web. Most of these images are regular, and for individual images, two lexicons are provided, including one 50-word lexicon and one 1000-word lexicon. 
\item \textbf{SVT} is a very challenging dataset collected by Wang et al.~\cite{ref21} from the Google Street View. Totally, 647 text images with low-resolution and noise are included. 
\item \textbf{CUTE} is released by Risnumawan et al.~\cite{ref24}. There are only 288 word images in this dataset, but most of them are seriously curved. Therefore, compared with other datasets, CUTE is more challenging. 	
\end{itemize} 

\subsection{Implementation Details}
 
In our experiments, all of the input images are scaled to a size of $64\times256$ with aspect ratio preserved. The maximal string length is set to 20, including one START token and one EOF token. This means up to 18 real characters are allowed within individual words. 
Our charset is composed of 39 characters, \textit{i.e.}, 26 alphabet letters, 10 digits, 1 START token, 1 EOS token and 1 special token for any other symbols.
The Adam optimizer with an initial learning rate of 1e-4  is employed in our work to optimize the proposed network. Totally, the proposed FACLSTM is trained for five epochs, with learning rates of 1e-4, 1e-4, 5e-5, 1e-5 and 1e-6, respectively. Moreover, the kernel size and channels ($N$ in Fig.~\ref{Figure5}) of the convolutional operations in Eqs.~\ref{equ4} and~\ref{equ5} are set to $3\times3$ and 256, respectively. Finally, the proposed network is implemented under the Tensorflow framework.

\subsection{Experimental Results}

We evaluate the performance of our proposed FACLSTM on the aforementioned three benchmark datasets, and compare it with those of the state-of-the-art approaches. Table~\ref{tab1} presents the details of the comparison results. Note that, in this table, CA-FCN~\cite{ref3} and SqueezedText~\cite{ref9} are the two latest recognizers recently published in AAAI2019 and AAAI2018.

\begin{table}[!t]
\footnotesize
\caption{Result comparison across different methods and datasets. Word-level recognition rate is used here. IIIT5K\_None, IIIT5K\_50 and IIIT5K\_1k denote no lexicon, 50-word lexicon and 1k-word lexicon are used, respectively. Samples: the number of samples used for training individual models, where * means datasets derived from SVT are used.}
\label{tab1}
\tabcolsep 8pt
\begin{tabular*}{\textwidth}{c|c|c|c|c|c|c|c}
\toprule
Method & LSTM & Samples & IIIT5K\_None & IIIT5K\_50 & IIIT5K\_1k & SVT & CUTE\\
\hline
{FAN~\cite{ref4}} & {FC-LSTM} & 12M* &{87.4} & {99.3} & {97.5} & {-}  & {63.9}\\

{AON~\cite{ref5}} & {FC-LSTM} & 12M* &{87.0} &  {\textbf{99.6}} & {98.1} &{-}  & {76.8}\\

{CRNN~\cite{ref6}} & {FC-LSTM} & 8M* &{78.2}  & {97.6} & {94.4}& {-}  & {-}\\

{(Gao et al.)~\cite{ref2}} & {FC-LSTM} & 8M* & {83.6} & {99.1} & {97.2} & {-}  & {-}\\

{RARE~\cite{ref7}} & {FC-LSTM} & 8M* & {81.9}  & {96.2} & {93.8}& {-}  & {59.2}\\

{R$^2$AM~\cite{ref25}} & {FC-LSTM} & 7M* & {78.4} &{96.8} &{94.4} &{-} &{-}\\

{SqueezedText\_binary~\cite{ref9}} & {FC-LSTM} & 1M & {86.6} & {96.9} & {94.3} & {-}  & {-}\\

{SqueezedText~\cite{ref9}} & {FC-LSTM} & 1M & {87.0}  & {97.0} & {94.1}& {-} &  {-}\\
\hline
{CA-FCN~\cite{ref3}} & {No} & 7M & {\textbf{92.0}} & {\textbf{99.8}} & {\textbf{98.9}}& {82.1}  & {\textbf{78.1}}\\

{(Gao et al.)~\cite{ref1}} &  {No} & 8M* & {81.8} & {99.1} & {97.9} & {-}  & {-}\\

{STN-OCR~\cite{ref8}} & {No} & - &  {86.0}  & {-} & {-}& {79.8} & {-}\\
\hline
{FLSTM\_base1} & {FC-LSTM} & 7M &{73.7} & {99.0} & {97.4} & {58.7} & {67.4}\\

{FAFLSTM\_base2} & {FC-LSTM} & 7M & {87.8} & {99.3} & {98.1} & {78.2} & {75.7}\\
\hline
{FACLSTM (proposed)} & {ConvLSTM} & 7M & {\textbf{90.5}} & {99.5} & {\textbf{98.6}} & {\textbf{82.2}} & {\textbf{83.3}}\\
\hline
\end{tabular*}
\end{table}

\textbf{Comparison with Methods based on the Traditional FC-LSTM:}
As previously introduced, traditional FC-LSTM is widely used in existing recognizers. 
Among the methods listed in Table~\ref{tab1}, RARE~\cite{ref7}, AON~\cite{ref5} and FAN~\cite{ref4} combined FC-LSTM with the attention mechanism in the fully connected way when performing sequential transcription, while CRNN~\cite{ref6}, R$^2$AM~\cite{ref25}, Gao's model~\cite{ref2} and SqueezedText~\cite{ref9} utilized FC-LSTM for frame-level prediction, sequential feature encoding or other purposes. 
As shown in Table~\ref{tab1}, our proposed FACLSTM outperforms these FC-LSTM-based methods by large margins on both regular text dataset IIIT5K (90.5\% vs 87.4\%) and curved text dataset CUTE (83.33\% and 76.8\%) when no lexicon is used. It also achieves competitive performance on IIIT5K when 1k-word lexicon and 50-word lexicon are used. Apparently, handling the text recognition task from the spatiotemporal perspective with our ConvLSTM-based FACLSTM is more effective than casting it to a sequence-to-sequence prediction problem via FC-LSTM, no matter for regular or irregular text images. Note that our FACLSTM is optimized with less training samples than most of the listed FC-LSTM-based recognizers, except for R$^2$AM~\cite{ref25} and SqueezedText~\cite{ref9}, and though AON~\cite{ref5} is specially designed for irregular text recognition, its recognition performance on CUTE is still 6.5\% lower than that of our FACLSTM.

Readers should keep in mind that apart from the 4 million training images from SynthText, the recognizers named AON~\cite{ref5} and FAN~\cite{ref4} also employed additional 8 million images provided by Jaderberg et al.~\cite{ref39} for their training. Jaderberg's synthetic images are generated with a 50k-word lexicon that covers all the test words of ICDAR and SVT datasets, and blended with word images randomly-sampled from these two datasets. Thus, the recognition performance on SVT would benefit largely from the usage of Jaderberg's images because of this strong correlation. This is also proved by Liao's work~\cite{ref3}, where a 4.3\% accuracy improvement on SVT was achieved by their CA-FCN when additional 4 million images generated with Jaderberg's strategy were used. In this work, to demonstrate the generalizability and robustness of proposed FACLSTM, we only employ the SynthText dataset to train our network. Therefore, to give a fair comparison, we only compare FACLSTM with recognizers not utilizing SVT-derived training images, such as CA-FCN~\cite{ref3} and STN-OCR~\cite{ref8}.

\textbf{Comparison with Non-LSTM based Methods:} Considering the limitations of the traditional FC-LSTM on neglecting spatial and structural information and slow training convergence etc, CA-FCN~\cite{ref3}, Gao's model~\cite{ref1} and STN-OCR~\cite{ref8} have also explored other non-LSTM solutions. Especially, CA-FCN~\cite{ref3} also addressed the recognition issue from the 2-D perspective by utilizing a FCN structure, and moreover, it used the same VGG-16 backbone and 7-million training images as our FACLSTM.

From Table~\ref{tab1}, we can see that the accuracy of our proposed FACLSTM is 1.5\% lower than that of the best recognizer, \textit{i.e.} CA-FCN~\cite{ref3}, on the regular text dataset IIIT5K. However, on the more challenging curved text dataset CUTE, we achieve an accuracy of 83.3\%, which is 5.2\% higher than that of CA-FCN~\cite{ref3}. As for the low-resolution and noisy dataset SVT, our FACLSTM performs slightly better than CA-FCN~\cite{ref3} with an accuracy of 82.2\% (vs. 82.1\% of CA-FCN~\cite{ref3}).   
Note that, CA-FCN~\cite{ref3} is not an end-to-end trainable system because in order to infer the final sequential outputs from the pixel-level predictions generated by their network, an empirical rule-based word formation module is required. By contrast, our FACLSTM is able to directly produce the final sequential outputs via the proposed ConvLSTM-based sequential transcription module. Admittedly, replacing FC-LSTM with Conv-LSTM will increase the memory cost. Therefore, to retain the efficiency, we up-sample feature maps to a small resolution of $1/4$ in the decoder branches, instead of $1/2$ used in CA-FCN. Undoubtedly, this small resolution will compromise the recognition accuracy to some extent, especially for small-size and low-resolution images from the IIIT5K and SVT datasets.

\textbf{Effectiveness of the Proposed Focused Attention Module and ConvLSTM-based Sequential Transcription Module:}
Furthermore, to highlight the effectiveness of our proposed focused attention module and ConvLSTM-based sequential transcription module, we compare the performance of our proposed FACLSTM with that of the following two baseline models:
\begin{itemize}
	\item FLSTM\_base1, which shares the same feature extraction module with our proposed FACLSTM, but removes the focused attention module. Besides, the sequential transcription module used in this model is the traditional attention-based FC-LSTM network, just as the one used in AON~\cite{ref5}, FAN~\cite{ref4} and both Gao's models~\cite{ref1, ref2}.  
	\item FAFLSTM\_base2, which is built upon FLSTM\_base1, but with the proposed focused attention module applied.
\end{itemize}

\begin{figure}[t]
\centering
\includegraphics[width=4in]{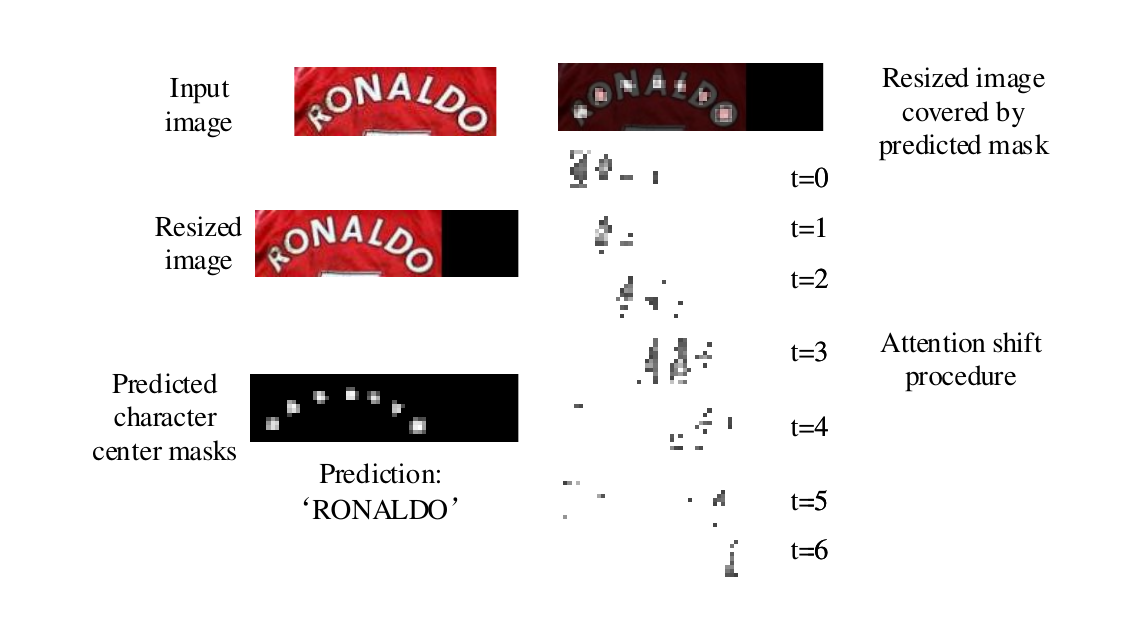}
\caption{Visualization results of predicted mask and attention shift procedure.}
\label{Figure6}
\end{figure}

Apparently, from the comparison of FLSTM\_base1 and FAFLSTM\_base2, we can see that the recognition accuracies on IIIT5K, SVT and CUTE datasets are elevated by 14.1\%, 19.5\% and 8.4\%, respectively when the proposed focused attention module is assembled. As illustrated in Figure.~\ref{Figure6}, the focused attention module is able to accurately predict the character center masks since it is performed in the high-level decoder branch. The significant performance improvement demonstrates that these masks are effective to help the sequential transcription module focus attention on the right character areas and suppress irrelevant background pixels. In addition, the image resolution of CUTE in much higher than that of SVT and IIIT5K, and SVT is much noisier than the other two datasets. As claimed in~\cite{ref3,ref4}, the attention-based recognizers perform poorly on low-quality images because of the `attention drift' problem, and the scene text images suffer from noisy background badly, so the accuracy improvement is more on SVT and less on CUTE when the proposed focused attention module is utilized.

Moreover, from the comparison of FAFLSTM\_base2 and FACLSTM, we can see that when the traditional attention-based FC-LSTM module is replaced by our proposed attention-ConvLSTM-based sequential transcription module, further 2.7\%, 3.6\% and 7.6\% improvements are achieved on IIIT5K, SVT and CUTE, respectively. This means our FACLSTM is able to boost the recognition performance significantly by utilizing the proposed attention-ConvLSTM module to take benefits from the valuable spatial and structural information of text images. As clarified in~\cite{ref3}, FC-LSTM only achieves good performance on horizontal or nearly horizontal text, and its performance on curved text is seriously limited because of the neglect of pixels' spatial correlation relationships. The huge performance improvement achieved by FACLSTM on CUTE evidences that our attention-ConvLSTM module is a good solution to this problem.

 Therefore, we can say that both of the proposed focused attention module and attention-ConvLSTM module are effective.
Note that the focused attention module can be removed from the network when datasets without character-level bounding box annotations are used for the training.

\begin{figure*}[t]
\centering
\includegraphics[width=5.8in]{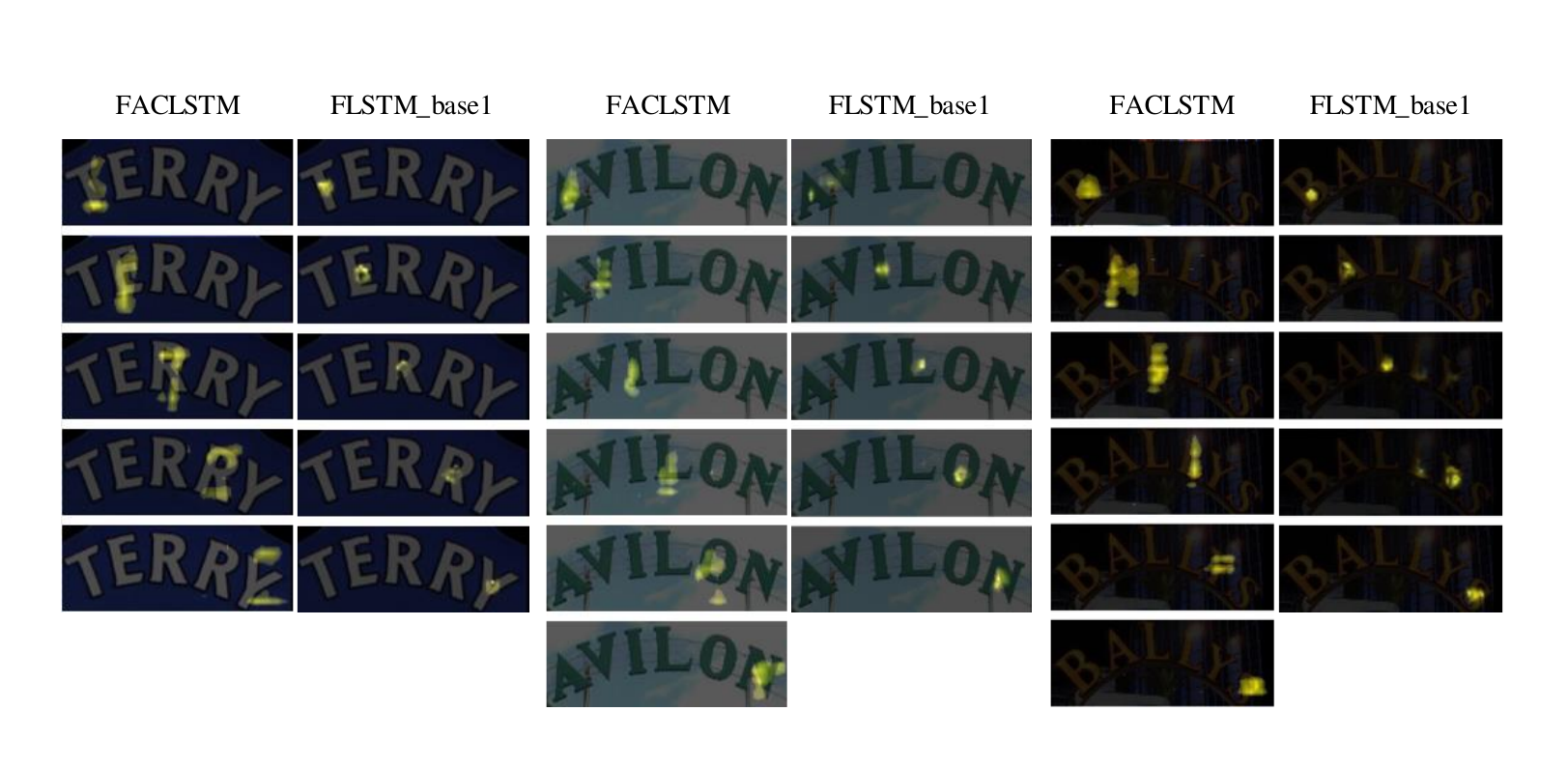}
\caption{Visualization results of attention predicted by FACLSTM and FLSTM\_base1. Values of the attention maps are normalized and truncated for a better visualization. Note that FACLSTM directly produces 2-D attention maps, while FLSTM\_base1 generates 1-D attention vectors, which are then reshaped to 2-D space.}
\label{Figure7}
\end{figure*}

In summary, on the regular text dataset, our proposed FACLSTM outperforms all of listed FC-LSTM-based and non-LSTM-based recognizers, except CA-FCN, but on the more challenging curved text dataset, our FACLSTM surpasses all of the listed methods significantly with an accuracy of 83.3\%, including CA-FCN (78.1\%). Moreover, the comparisons with other two baseline models demonstrate the effectiveness of our proposed focused attention module and ConvLSTM-based sequential transcription module. Finally, we also give the visualization results of the predicted masks and the attention shift procedure, as shown in Fig.~\ref{Figure6}. The comparison results of attention predicted by FACLSTM and FLSTM\_base1 are shown in Fig.~\ref{Figure7}. Note that our FACLSTM directly produces 2-D attention maps via the convolutional operations, while FLSTM\_base1 generates 1-D attention vectors with the fully connected layers, just as other existing FC-LSTM-based recognizers did. These 1-D attention vectors are reshaped to 2-D maps in Fig.~\ref{Figure7} for an intuitional visualization. As we can see, the attention areas of FACLSTM is larger and more accurate, and the `attention drift' problem is alleviated to some extent in our proposed FACLSTM.

\section{Conclusion}
\label{sec:con}
Scene text recognition has been treated as a sequence-to-sequence prediction problem for quite a long time, and traditional FC-LSTM is widely used in current state-of-the-art recognizers. 
In this work, we have demonstrated that scene text recognition is actually a spatiotemporal prediction problem and we have proposed to tackle this problem from the spatiotemporal perspective. Toward this end, we have presented an effective scene text recognizer named FACLSTM, where ConvLSTM was applied and improved by integrating the attention mechanism in the sequential transcription module, and a focused attention module has been designed at the encoder-decoder feature extraction stage.
Experimental results have revealed that, our proposed FACLSTM is able to handle both regular and irregular (low-resolution, noisy and curved) text well. Especially for the curved text, our proposed FACLSTM has outperformed other advanced approaches by large margins. 
Thus, we can conclude that ConvLSTM is more effective in scene text recognition than the widely used FC-LSTM since the valuable spatial and structural information can be better leveraged when performing sequential prediction with ConvLSTM.


\Acknowledgements{This work was supported by China Scholarship Council (No. 201706140138), Shanghai Natural Science Foundation (No. 19ZR1415900) and Shanghai Knowledge Service Platform Project (No. ZF1213).}




\end{document}